\documentclass{article}

\usepackage[preprint,nonatbib]{neurips_2022}

\usepackage[utf8]{inputenc} 
\usepackage[T1]{fontenc}    
\usepackage{hyperref}       
\usepackage{url}            
\usepackage{booktabs}       
\usepackage{amsfonts}       
\usepackage{nicefrac}       
\usepackage{microtype}      
\usepackage{xcolor}         

\usepackage{multirow}
\usepackage{amsmath,amssymb,amsthm}
\usepackage{graphicx}
\usepackage{subcaption}
\usepackage{mathtools}
\usepackage{bm}
\usepackage{diagbox}
\usepackage{wrapfig}
\usepackage{algorithm}
\usepackage{algorithmic}
\usepackage{makecell}
\usepackage{longtable}
\usepackage{cases}

\theoremstyle{plain}

\theoremstyle{definition}

\theoremstyle{remark}

\title{A Game-Theoretic Perspective of Generalization in Reinforcement Learning}

\author{%
  Chang Yang\thanks{Equal contribution.} \\
  National University of Singapore\\
  \texttt{yangchang@u.nus.edu} \\
   \And
  Ruiyu Wang\footnotemark[1] \\
  National University of Singapore\\
  \texttt{ruiyu.wang@u.nus.edu} \\
  \And
  Xinrun Wang\thanks{Corresponding author.} \\
  Nanyang Technological University\\
  \texttt{xinrun.wang@ntu.edu.sg} \\
  \And
  Zhen Wang\\
  Hangzhou Dianzi University\\
  \texttt{wangzhen@hdu.edu.cn}
}

\begin{document}

\maketitle
\begin{abstract}
Generalization in reinforcement learning (RL) is of importance for real deployment of RL algorithms. Various schemes are proposed to address the generalization issues, including transfer learning, multi-task learning, meta learning, as well as robust and adversarial reinforcement learning. However, there is not a unified formulation of
various schemes and comprehensive comparisons of methods across different schemes. In this work, we propound GiRL, a game-theoretic framework for generalization in reinforcement learning, where a RL agent is trained against an adversary over a set of tasks, over which the adversary can manipulate the distributions within a given threshold. With different configurations, GiRL is capable of reducing the various schemes mentioned above. To solve GiRL, we adapt the widely-used method in game theory, policy space response oracle (PSRO) framework with three significant modifications as follows: i) we adopt model-agnostic meta learning (MAML) as the best-response oracle, ii) we propose a modified projected replicated dynamics, i.e., R-PRD, which ensures the computed meta-strategy for the adversary falls in the threshold, and iii) we also propose a protocol of few-shot learning for multiple strategies during testing. Extensive experiments on MuJoCo environments demonstrate that our proposed method outperforms state-of-the-art baselines, e.g., MAML.
\end{abstract}

\section{Introduction}
In terms of real world deployment of reinforcement learning (RL) algorithms, it's critical for the RL models to perform robustly on unseen test scenarios. Various schemes are proposed to tackle this generalization issues~\cite{kirk2021survey}, including transfer learning, multi-task learning, meta learning, robust and adversarial reinforcement learning. Transfer learning focus on transferring the policy to a new task without seeing before~\cite{pan2009survey}, multi-task learning intends to learning a policy which can perform well on several tasks~\cite{ruder2017overview}, while meta learning, i.e., ''learning to learn'',  tackles the cases of quickly adapting a trained policy to new tasks~\cite{finn2017model}. Different from previous scheme, robust and adversarial reinforcement learning consider to improve the agent's performance under attacks from the adversary or uncertainties from the real world~\cite{pinto2017robust}. The considered attacks 
include attacking on both the agent's body and observations. Though different learning schemes are able to handle generalization issues, they are often regarded as different research fields. Therefore, there is not a unified formulation of generalization in reinforcement learning and their methods are not well compared across different fields. 

To narrow the gaps across different research fields, in this work, we leverage the game-theoretic methodology to provide a unified perspective of the generalization in reinforcement learning, which we term as \textbf{GiRL}. Specifically, GiRL considers the generalization as the game played between an RL agent and an adversary, where the agent learns to perform well on a set of tasks while the adversary is to manipulate the distribution (within a threshold) over tasks to decrease the agent's performance. The number of tasks can be finite as in multi-task learning, or infinite as in robust learning and with different configurations, GiRL can reduce the different schemes mentioned above. 

To construct GiRL, we adapt the widely-used policy space response oracle (PSRO) in game theory. PSRO starts with a set of randomly initialized policies, i.e., RL agents, and iteratively add new policies into consideration through the best-response oracle, which computes the best-response of the agent against a set of tasks under a specific distribution. We make three critical modifications to PSRO for GiRL: i) we use model-agnostic meta learning (MAML) as the best-response oracle, rather than the naive RL algorithms, e.g., PPO, where MAML demonstrates its superior performances against a set of tasks, ii) we propose a modified projected replicated dynamics (PRD) which ensures the computed meta-strategy of the adversary fall in the threshold (which is also used during the evaluation), and iii) we also propose a protocol for the few-shot learning of the multiple strategies during testing. Extensive experiments are performed on multiple MuJoCo environments, such as AntVel and AntPos to demonstrate that our proposed methods can outperform existing baselines.

\section{Related Works}
In this section, we present a brief overview of related works. There are mainly four lines of research related to our works, reinforcement learning, multi-task/transfer/meta learning, robust and adversarial reinforcement learning and game-theoretic methods for machine learning.

\paragraph{Reinforcement Learning.} Deep reinforcement learning (DRL) is widely applied to solve complex decision-making tasks. Most of the DRL methods depend on the temporal-difference learning, which learns the state-vale function by bootstrapping.  Examples of off-policy DRL methods, where policies are trained under data collected by other policy, include DQN~\cite{mnih2015human}, A3C~\cite{mnih2016asynchronous} and SAC~\cite{haarnoja2018soft}. While there are only two widely used on-policy DRL methods, i.e., the policy is trained under data collected by itself: TRPO~\cite{schulman2015trust} and PPO~\cite{schulman2017proximal}, among which PPO is developed as a refinement of TRPO and both methods rely on the usages of trust regions. TRPO serves as the backbone algorithm in our experiments. 

\paragraph{Multi-task/Transfer/Meta Learning.} The three schemes, multi-task learning, transfer learning and meta learning, are often regarded as related fields, while there are some slightly differences in their configurations. Multi-task learning focuses on learning an agent which can perform well across different tasks, transfer learning is that training an agent on one task and transfer the agent's policy to another task, while meta learning focus on improving the learning on new tasks with the knowledge learned from existing tasks. Various methods include policy distillation, i.e., distilling the knowledge of an expert or a set of experts to a student policy via knowledge distillation~\cite{hinton2015distilling,rusu2016policy,parisotto2016actor} and representation learning methods that transfer the learned feature representations to facilitate the policy learning~\cite{devin2017learning,barreto2018transfer}. We refer~\cite{zhu2020transfer,ghosh2021generalization,kirk2021survey} for detail reviews of these methods. Among the various methods proposed in the literature, Model-Agnostic Meta Learning (MAML) is one of the most notable methods, which demonstrates its superior performance on various tasks~\cite{finn2017model,raghu2019rapid}. In this work, we provide a game-theoretic perspective of these learning schemes, where an adversary is introduced to manipulate the task distribution and with different assumptions of the adversary, we can obtain the different schemes. 


\paragraph{Robust and Adversarial Reinforcement Learning.} Robustness is another important measure for the generalization, which tries to achieve good performance under uncertainties or adversarial attacks. And it becomes even more critical when the policy is represented by neural networks, because neural networks are vulnerable to small perturbations~\cite{pinto2017robust}. One of the earliest work to investigate the robustness in deep RL is~\cite{pinto2017robust}, in which an adversary is introduced into training, who 
has the power to attack the body of the RL agent. They mainly follows the adversarial training process, where the RL agent and the adversary are trained alternatively. Subsequent works include increasing the diversity of the policies~\cite{vinitsky2020robust,gleave2019adversarial} and considering different perturbations by the adversary~\cite{zhang2020robust}. However, the training of RARL is often unstable because a strong adversary will largely decrease the agent's performance and make the training unsuccessful~\cite{zhang2020stability}. Therefore, we can leverage game-theoretical methods to stabilize the training. Furthermore, instead of only considering the specific attacks, introducing an adversary which can manipulate the task distribution can provide a more general framework for the generalization in RL. 

\paragraph{Game-theoretic Methods for Machine Learning.} Game theory provides a natural framework to model the interactions between two competitive players, such as poker~\cite{brown2018superhuman}, Go~\cite{silver2016mastering} and even StarCraft II~\cite{vinyals2019grandmaster}. A simple yet powerful framework, policy space response oracle (PSRO),~\cite{lanctot2017unified,muller2020generalized} extends the widely-used double oracle (DO) method~\cite{mcmahan2003planning,jain2011double} to deep version where a RL-based oracle is used to compute the best-response because the computing of best-response is reduced to an optimization problem when the opponent is fixed. The basic idea of PSRO is to start with a set of heuristic or randomly policies and iteratively add the best-responses into the policy set. The main advantage of PSRO is that it can approximate the equilibrium solution without explicitly enumerating all policies. Thus, PSRO can also be applied to game with continuous actions. PSRO as well as DO framework are used to solve large-scale normal-form and extensive-form games such as security games~\cite{tsai2012security,jain2011double}, poker games~\cite{waugh2009strategy} and search games~\cite{bosansky2012iterative} and they are widely used in various disciplines. Recently, DO is successfully applied to generative adversarial networks (GAN)~\cite{aung2021gan}, which demonstrates the potential of game-theoretic methods in improving the performance of traditional machine learning methods on problems with competitive players. In this paper, we exploit the framework of PSRO for GiRL with modifications in the best-response oracle and meta-solver.

\section{Preliminaries}
In this section, we present the preliminaries of game theory, Markov Decision Process (MDP) and the formulation of GiRL.

\paragraph{Game Theory.} A normal-form game is a tuple $(\mathcal{A}, U, N)$ where $N = \{1, \dots , n\}$ is a set of players, $\mathcal{A} = \prod_{i=1}^{n}\mathcal{A}_{i}$ is a pure strategy space with pure strategies $\mathcal{A}_{i}$ for each player $i \in N$ and $U: \mathcal{A} \rightarrow R^{n}$ is a payoff table for each joint policy played by all players. Each player chooses strategy among $\mathcal{A}_{i}$ to maximize own expected utility, or by sampling from a mixed strategy distribution $\pi_{i} \in \Delta(\mathcal{A}_{i})$, where $\bm{\pi}$ is known as the mixed strategy profile. The utility of the player $i$ under mixed strategy profile $\bm{\pi}$ gives $u_{i}(\bm{\pi})=\sum_{\bm{a}\in\mathcal{A}}U_{i}(\bm{a})\prod \limits_{j=1}^n\pi_{j}(a_{j})$. The canonical solution concept in game theory is Nash Equilibrium (NE), in which no player can unilaterally deviate from the strategy to increase his utility: a mixed strategy profile $\bm{\pi}$ forms an NE if
\begin{equation}
    u_{i}(\bm{\pi})\geq u_{i}(\bm{\pi}_{-i}, \pi'), \forall i\in N, \pi'\in\Delta(\mathcal{A}_{i})
\end{equation}
Computing NE in general-sum games is PPAD-Complete~\cite{daskalakis2009complexity}, and even computing the NE in two-player zero-sum games is non-trivial. Therefore, DO framework and its deep version PSRO are proposed to compute the NE in games with large action and state spaces and even continuous action and state spaces, in which case instead of enumerating all actions, DO iteratively adds new best-responses into the restricted games and PSRO utilizes RL methods to compute new best-responses.

\paragraph{Markov Decision Process (MDP).} In RL, a task is often represented by a Markov decision process (MDP). A MDP is defined by a tuple: $\left(\mathcal{S}, \mathcal{A}, \mathcal{P}, r, \gamma, T\right)$, where $\mathcal{S}=\langle s\rangle$ is the state space and $\mathcal{A}=\langle a\rangle$ is the action space. $\mathcal{P}: \mathcal{S} \times \mathcal{A} \times \mathcal{S} \rightarrow [0, 1]$ is the transition function which specifies that the probability of reaching $s'\in\mathcal{S}$ at the next state, when taking $a\in\mathcal{A}$ at the current $s\in\mathcal{S}$, $r: \mathcal{S} \times \mathcal{A} \times \mathcal{S}\rightarrow \mathbb{R}$ is the reward function which specifies the reward of the agent when taking $a\in\mathcal{A}$ at $s\in\mathcal{S}$ and reach $s'\in\mathcal{S}$, $\gamma \in(0,1]$ is the discount factor and $T$ is the time horizon\footnote{When $\gamma<1$, $T$ can be $\infty$.}. In a MDP, agent receives current state $s_{t} \in \mathcal{S}$ from the environment and performs an action $a_{t} \in \mathcal{A}$ according to the policy $\pi_{\theta}: \mathcal{S}\times\mathcal{A}\rightarrow [0, 1]$, which is parameterized by $\theta$. The objective of the agent is to learn an optimal policy $\pi^{*}:=\arg\max_{\theta} \mathbb{E}_{\pi_{\theta}}\left[\sum_{t=0}^{T} \gamma^{t} r_{t} | s_{0}\right]$, where $s_{0}$ is the initial state of the MDP. 


\paragraph{Contextual MDP.} We follow \cite{kirk2021survey} to use the contextual Markov decision process ($\mathbb{C}$MDP) as our formalism of GiRL. Specifically, a $\mathbb{C}$MDP $\mathcal{M}$ is a MDP whose states can be decomposed into tuples $s=(c, s')$, where $s'\in\mathcal{S}$ is the underlying state and $c\in\mathcal{C}$ is referred as \emph{context} for the state which is constant during an episode\footnote{$c$ is the index of the context when $\mathcal{C}$ is discrete and a context vector when $\mathcal{C}$ is continuous.}. Each context corresponds to a different situation that the agent might be in, with slightly difference in dynamics and rewards, and some shared structure across which an agent can generalize. During training, the agent has access to a subset of contexts $\mathcal{C}_{train}\subseteq \mathcal{C}$ with a known distribution $p_{0}$ over $\mathcal{C}_{train}$\footnote{We note that a distribution over $\mathcal{C}_{train}$ is also a distribution over $\mathcal{C}$.}. When testing, we consider that there is an adversary who manipulates the test distribution $p_{1}\in \delta(\mathcal{C})$ over all contexts for evaluation. We assume that $\delta(\mathcal{C})=\{p~|~D(p, p_{0})\leq\epsilon\}$, where $D(\cdot, \cdot)$ is the distance between the two distributions, e.g., Kullback–Leibler (KL) divergence, and $\epsilon$ is the perturbation budget. During testing, we conduct $K$-shot ($K\geq 1$) evaluation for the agent, so that average performance is estimated over $K$-shot or the last shot. The evaluation vector $\bm{\alpha}$ tracks the weights for few-shot evaluation. We intend to train a set of policies $\mathcal{F}=\{f_{i}\}$ for the agent, where $f$ is a specific policy maps the observation to the action. $\pi$ is the distribution over $\mathcal{F}$. We assume that $|\mathcal{F}|\leq M$ where if $M=1$, there is only one policy allowed as in meta learning. The performance of the agent according to the evaluation protocol is denoted as $U(c, \mathcal{F})$ under context $c$. We assume that the adversary intends to decrease the performance of the agent, while the adversary can be bounded rational: a simple way to handle the bounded rationality is to introduce a smooth parameter $\beta$ between $p_{0}$ and the optimal distribution $p_{1}^{*}=\arg\min_{p_{1}'\in\delta(\mathcal{C})}p_{1}'(c)U(c,\mathcal{F})$, i.e., the distribution for the adversary is $p_{1} = \beta p_{0} + (1-\beta)p_{1}^{*}$. Our goal is to maximize the performance of the agent during testing, i.e., $\int_{c\in\mathcal{C}}p_{1}(c)U(c,\mathcal{F})$.

\begin{table}[ht]
\centering
\caption{Summary of the types of learning. For those types of learning with $\epsilon=0$, we can also add a small perturbation, e.g., $1e-5$ to check whether the training can improve the performance.}
\label{tab:summary}
\begin{tabular}{l||ccccc}
\toprule
Types of Learning& $\mathcal{C}_{train}$ & $\epsilon$ & $K$ & $\bm{\alpha}$ & $\beta$\\
\midrule
Robust learning     & $\mathcal{C}$ &$>0$&$1$&$[1]$& $0$\\
Adversarial learning     & $\mathcal{C}$&$>0$&$1$&$[1]$& $0$\\
Multi-task learning & $\mathcal{C}$ &$0$&$1$&$[1]$& $1$\\
Meta learning & $<\mathcal{C}$ &$0$&$>1$&$[0,\dots, 1]$& $1$\\
Transfer learning & $<\mathcal{C}$ &$>0$&$1$&$[1]$& $1$\\
\bottomrule
\end{tabular}
\end{table}
\paragraph{From GiRL to Various Learning Schemes.} We summarize the relations of the different configurations and various schemes mentioned above in Table~\ref{tab:summary}. Specifically, the adversary is fully random in multi-task/meta/transfer learning, because the training distribution and the test distribution over tasks are the same and the agent can maximize the sum of the utilities obtained from different tasks, and the adversary is fully rational in robust and adversarial reinforcement learning, where the test distribution is intentionally selected to decrease the agent's performance. The agent can access to all the task in robust, adversarial, multi-task learning and can access partial of the tasks in meta and transfer learning. The distribution shift of the task is considered in robust, adversarial and transfer learning, and ignored in multi-task and meta learning. We note that different from the multi-task/meta/transfer learning, robust and adversarial reinforcement learning do not explicit have the set of the tasks, because the specific task that the agent need to complete is determined by the adversary policy. However, we can regard the set of tasks as tasks induced by all possible adversary's policies. With different configurations of GiRL, we can easily reduce it to different learning scheme. Therefore, GiRL enables us to handle various learning schemes in a unified framework and even tackle configurations which are not considered in current literature. Thus, GiRL opens new venues for generalization in reinforcement learning.

\section{The Proposed Method}
\label{sec:psro_girl}
GiRL formulates the generalization in reinforcement learning as a game between two players, the agent and the adversary. To compute the optimal policies of both players, we leverage the widely-used framework, policy space response oracle (PSRO), for GiRL, i.e., PSRO-GiRL, which is a modified PSRO framework that can handle the following issues: i) deriving the best-response against a set of tasks with MAML rather than the naive RL methods, ii) computing the meta-strategies in a restricted strategy space with a restricted projected replicator dynamics (R-PRD), and iii) adopting few-shot learning of the learned strategies during evaluation. We also design our evaluation protocol for fair comparisons. We note that though GiRL provides a general framework for RL generalization, in this paper, we only focus on the case with finite number of tasks, agent that can access all tasks, i.e., $\mathcal{C}_{train}=\mathcal{C}$ and rational adversary, i.e., $\beta=1$. The details will be presented in the rest of this section. 
\begin{figure}[ht]
\centering
\includegraphics[width=\textwidth]{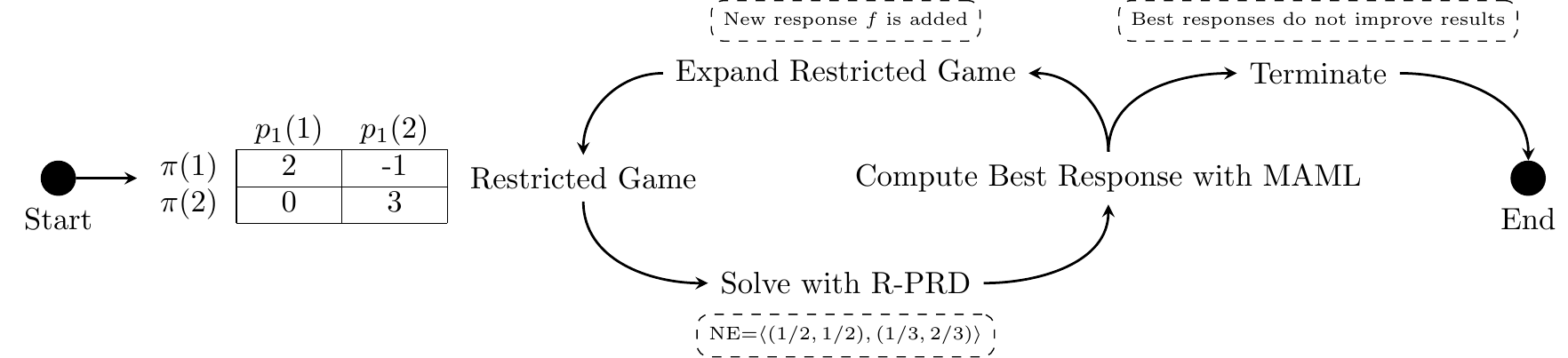}
\caption{An illustration of the flow of PSRO-GiRL. Figure adapted from~\cite{lanctot2017unified}. }
\label{fig:my_label}
\end{figure}

\begin{algorithm}
\caption{Policy space response oracle (PSRO) for GiRL}
\label{alg:psro_girl}
\begin{algorithmic}[1]
\STATE Initialize the uniform distribution $p_{1}$ over tasks as in MAML, an empty policy set $\mathcal{F}$, the meta-strategy distribution $\pi$ for the agent, and the meta-game payoff table $U$\label{alg:init}
\WHILE{$loop\in\{1, 2, \dots\}$}
\STATE Compute the agent's best-response $f$ with MAML against $p_{1}$ and $\mathcal{F}=\mathcal{F}\cup\{f\}$ \label{alg:best_response}
\STATE Augment the meta game with the new response $f$ to obtain the updated $U$ \label{alg:augment}
\STATE Compute the mixed strategies of the agent and the distribution over tasks $\langle \pi, p_{1}\rangle$\label{alg:meta}
\ENDWHILE
\end{algorithmic}
\end{algorithm}
\paragraph{Policy Space Response Oracle (PSRO).} PSRO is proposed for computing NE in games with large-scale, even continuous, action and state spaces. We present the general proceduce of PSRO-GiRL in Algorithm~\ref{alg:psro_girl}. The normal PSRO starts with the randomly initialized policies. However, in this work, we consider the case with finite number of tasks. Therefore, we can initialize the task distribution $p_{1}$ as a uniform distribution (Line~\ref{alg:init}). Then, we can compute the best-response of the agent. The normal PSRO uses RL methods, e.g., PPO, to compute the optimal policy against a set of adversary's policies, which corresponds to the distribution over the tasks. However, We use MAML as our best-response oracle (Line~\ref{alg:best_response}), which demonstrates its superiority against a set of tasks (which is detailed below). After obtain the best-response $f$, we evaluate the new policy on the tasks and augment the meta-game payoff table $U$ (Line~\ref{alg:augment}). Then, we apply the restricted projected replicator dynamics (R-PRD) to compute the meta-strategies, i.e., the distributions over the responses computed $\pi$ and the distribution over tasks $p_{1}$ (Line~\ref{alg:meta}). Then the algorithm will terminated if the max loop is reached, otherwise a new loop will be started. The current PSRO framework can be easily extended to include i) more criteria for determining the termination, ii) more solution concepts to be considered, e.g., $\alpha$-rank distribution~\cite{muller2020generalized}, and ii) more advanced response oracle, e.g., ANIL~\cite{raghu2019rapid}. More advanced techniques will be considered in future works.

\begin{algorithm}
\caption{Model Agnostic Meta Learning (MAML) as the Response Oracle}
\label{alg:maml}
\begin{algorithmic}[1]
\STATE Given the distribution $p_{1}$ over tasks, a policy $f$ parameterized by $\theta$. 
\WHILE{Not terminated}
\STATE Sample a set of tasks $\mathcal{C}_{sample}$ from $\mathcal{C}_{train}$ according to $p_{1}$\label{alg:sample_tasks}
\FOR{task $c$ in $\mathcal{C}_{sample}$}
\STATE Sample $K$ trajectories $\mathcal{D}_{c}$ from $c$\label{alg:inner_start}
\STATE Evaluate compute the gradient $\nabla_{\theta}\mathcal{L}_{c}$ where $\mathcal{L}_{c}=-\mathbb{E}_{\mathcal{D}_{c}}[\sum_{t=0}^{T}\gamma^{t}r_{t}]$
\STATE Update the policy with the gradient and obtain $f'_{c}$: $\theta'_{c}=\theta-\alpha \nabla_{\theta}\mathcal{L}_{c}$\label{alg:inner_end}
\STATE Sample $K$ new trajectories $\mathcal{D}'_{c}$ with $f'_{c}$
\ENDFOR
\STATE Update the policy using $\mathcal{D}'_{c},\forall c\in\mathcal{C}_{sample}$: $\theta=\theta - \beta\nabla_{\theta}\sum\nolimits_{c\sim p_{1}(c)}\mathcal{L}_{c}(f'_{c})$\label{alg:outer_update}
\ENDWHILE
\end{algorithmic}
\end{algorithm}
\paragraph{MAML as the Response Oracle.} We choose MAML as the response oracle to compute the best-response of the agent at each loop, given that MAML outperforms naive RL methods, e.g., TRPO, in dealing with multiple tasks. We present the procedure of MAML in Algorithm~\ref{alg:maml}. With the given task distribution compute by previous loop $p_{1}$, computing the best-response is equivalent to finding a policy which can achieve good performances on multiple tasks. Specifically, MAML sample a set of tasks from the training set of tasks $\mathcal{C}_{sample}$ according to the distribution $p_{1}$ (Line~\ref{alg:sample_tasks}) and then for each sample tasks, MAML conducts the update of the policy to obtain a task-specified policy $f'_{c}$ (Lines~ \ref{alg:inner_start}-\ref{alg:inner_end}) and then use the new policy to sample the trajectories $\mathcal{D}'_{c}$. After the inner loop, all the sampled trajectories of the sampled tasks is used to update the policy $f$ (Line~\ref{alg:outer_update}). In the experiments, we use TRPO as the backbone of the MAML update. 


\paragraph{Meta-Solver for Restricted Strategy Space.} At each loop of PSRO-GiRL, we need to compute the strategies of the agent and the adversary. The traditional solution methods such as linear program can not be directly applied to our case because we introduce the constraints between the manipulated task distribution $p_{1}$ and the known distribution $p_{0}$. Therefore, novel meta-solver is needed. Inspiring by the projected replicator dynamics (PRD) proposed in~\cite{lanctot2017unified}, we propose restricted PRD (R-PRD) to handle the restriction of the strategy space. Specifically, given the payoff table $U$, for convenience we represent the payoff tables of the agent and the adversary as $A$ and $B$, respectively. The replicator dynamics~\cite{bloembergen2015evolutionary} update the strategies of both players $\langle\pi,p_{1}\rangle$ as:
\begin{align}
    \frac{d \pi(i)}{dt}&=\pi(i)[(A p_{1})_{i}-\pi^{\top}Ap_{1}] \label{eq:rd_agent}\\
    \frac{dp_{1}(j)}{dt}&=p_{1}(i)[(\pi^{\top}B)_{j}-\pi^{\top}Bp_{1}]\label{eq:rd_adversary}
\end{align}
where $\top$ means transpose. The PRD is proposed to ensure the exploration during the update, thus improving the convergence of the learning, where a projection operator $\text{Proj}(\cdot)$ is introduced to map the action with the probability lower than a threshold to the threshold~\cite{lanctot2017unified}, where $\text{Proj}(x)=\arg\min_{x'\in\delta^{\epsilon}}\{||x'-x||\}$ and $\delta^{\epsilon}=\{x|x(i)>\epsilon, \sum_{i}x(i)=1\}$. In our case, we need to ensure not only the exploration, but also the strategy of the adversary, i.e., distribution over tasks, fall in the set $\delta(\mathcal{C})$. Therefore, in this paper, we focus on the $\infty$-norm of the distribution, where we introduce the upper bound $\overline{p}$ and the lower bound $\underline{p}$ of the probability that a task can have. Therefore, after each update with PRD, we clip each element in the strategy $p_{1}$ with $[\overline{p}, \underline{p}]$ and normalize the sum of the elements into 1. We repeat this process until the $p_{1}$ falls in $\delta(\mathcal{C})$ (Line~\ref{alg:clip}). Then we use the projection operator to further refine the strategies of both players for exploration and start a new loop (Line~\ref{alg:proj}). The detailed procedure is presented in Algorithm~\ref{alg:psro_girl}.
\begin{algorithm}
\caption{Restricted Projected Replicator Dynamics (R-PRD) as meta solver} 
\begin{algorithmic}[1]
\STATE Given the current payoff table $U$, and the uniform distribution over policies and tasks $\langle\pi, p_{1}\rangle$
\WHILE{Not converged}
\STATE Update $\langle\pi, p_{1}\rangle$ with Eqs. (\ref{eq:rd_agent})-(\ref{eq:rd_adversary}).
\WHILE{$p_{1}\not\in\delta(\mathcal{C})$ }
\STATE $p_{1}(i)\leftarrow \text{CLIP}(p_{1}(i), \overline{p}, \underline{p}),\forall i$\label{alg:clip}
\ENDWHILE
\STATE $\pi\leftarrow \text{Proj}(\pi)$,\qquad $p_{1}\leftarrow \text{Proj}(p_{1})$\label{alg:proj}
\ENDWHILE
\end{algorithmic}
\end{algorithm}
\paragraph{Fine-Tuning during Testing.} 
The above parts focus on the training where no few-shot learning is considered. Different from MAML which are usually return a policy, GiRL returns a set of policies, e.g., 5. Therefore, for the few-shot learning, we assume that during the testing, all the policies are trained by the few-shot learning, which is mostly in line with the evaluation in game theory where the mixed strategies will be evaluated for enormous times to compute the expected utility. We can also consider that within the limited interaction, how to efficiently fine-tuning the multiple policies, which will be investigated in future works. And then the agent will use the same $\pi$ during training, i.e., we do not recompute the meta-strategies of the agent. During testing, we assume that the adversary can observe the agent's updated policies and find the optimal distribution $p_{1}$ which can decrease the performance of the agent, i.e., the adversary can update the meta-strategy during testing. The finding of the optimal $p_{1}$ is also using Algorithm~\ref{alg:psro_girl} with the fixed agent. We choose this as this is the worst case for the agent during testing. 

\section{Experimental Evaluation}
\subsection{Experiment Setup}



To evaluate PSRO-GiRL on RL problems, we consider two widely-used tasks, Ant Velocity and Ant Position, in MuJoCo environment. We design experiments on these sets of tasks, and compare the performance of PSRO-GiRL against the baseline MAML. 
Each of the model trained by PSRO-GiRL or MAML is a neural network policy, with two hidden layers of size 100, and with ReLU activation. We use trust-region policy optimization (TRPO) as the meta-optimizer, which is an on-policy RL method based on trust regions and is widely used in meta-learning. To be mentioned, PSRO-GiRL consists of two cases, i.e., *GiRL and GiRL, in which *GiRL we reinitialize the policy network at each iteration, while in GiRL we continue to train the policy network without reinitialization.  All algorithms are implemented by PyTorch~\cite{paszke2019pytorch} and all experiments are performed on a 64-bit machine with 56 Intel(R) Xeon(R) CPU E5-2683 v3 CPUs and 4 NVIDIA Tesla V100 GPUs.

\paragraph{Evaluation Protocol.} Given that PSRO-GiRL has multiple policies, while MAML only has one policy, we leverage the following protocol for fair comparisons: Consider that we have $M$ policies, where $M=1$ in MAML and $M>1$, e.g., 5, in PSRO-GiRL (policies with positive probabilities in the final equilibrium strategy). Assume we have $N$ tasks, and conduct $K$-shot learning, where $K=\{0, 1, 3, 5\}$, of all policies on these tasks, and then evaluate the policies. Therefore, we obtain $M\times N$ matrices, i.e., $1\times N$ in MAML and $5\times N$ in PSRO-GiRL. Then, we multiply the matrix with the policy distribution, so both MAML and PSRO-GiRL will get $1\times N$ reward matrix $R$. Given that the train distribution over environments is $p_{0}$, we let the adversary manipulate the distribution, computed by Algorithm~\ref{alg:meta}, where the agent is fixed, against the reward matrices for both MAML and PSRO-GiRL, i.e., $p_{1}\cdot R$. Therefore, we select the worst distribution over tasks $p_{1}$ to evaluate the agent, then calculate the average returns for both MAML and PSRO-GiRL.
\subsection{Results}
\begin{wrapfigure}{r}{0.35\textwidth}
\begin{center}
\includegraphics[width=0.35\textwidth]{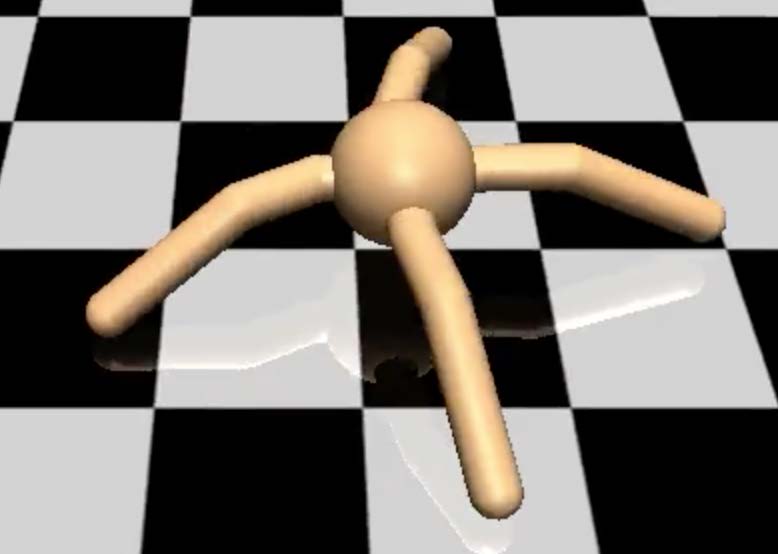}
\end{center}
\caption{Ant environment. }\label{fig:ant}
\end{wrapfigure}
To study the validity of PSRO-GiRL for complex deep RL problems, we adopt ant locomotion(including Ant Velocity and Ant Position) as our meta-RL experiment and generate a set of high-dimensional locomotion tasks with MuJoCo simulator. In training process, we set hyper-parameters $train\_max$ and $train\_min$ as upper and lower limits for the probability distribution of the tasks to make sure the probability corresponding to each task is within the range controlled by the upper and lower limits, while the sum of the probabilities of all tasks always equals to 1. Three runs were carried out under each setup with 3 different seeds. In testing process, $test\_max$ and $test\_min$ similarly serve as hyper-parameters for upper and lower limits. The saved model were fine-tuned on its corresponding task for $K$ times ($K$-shot) before evaluation, where $K$ is a hyper-parameter. The average and standard deviation values of each run are calculated.

\begin{table}[ht]
\centering
\caption{Experiments on AntVel with 20 diverse tasks.}
\label{tab:0_shot_antvel}
\begin{tabular}{c|c|c||c|cc}
\toprule\toprule
Shots&Test max     & Train max & MAML & *GiRL (Ours)& GiRL (Ours) \\
\midrule
\multirow{9}{*}{0-shot}&\multirow{3}{*}{0.1}& 0.1 &  \multirow{3}{*}{-46.23 (3.31)}& -32.11 (6.19) & -28.84 (6.47)\\ 
&& 0.2 &  & -27.89 (6.09) & -24.76 (12.44) \\
&& 0.3 &  & -32.71 (7.73) & -21.71 (2.90)\\
\cmidrule{2-6}
&\multirow{3}{*}{0.2}& 0.1 &  \multirow{3}{*}{-67.50 (5.15)}&  -42.41 (10.42) & -39.13 (10.86)\\
&& 0.2 &  & -36.84 (9.74) & -33.00 (18.13) \\
&& 0.3 &  & -42.76 (8.47) & -28.04 (2.24)\\
\cmidrule{2-6}
&\multirow{3}{*}{0.3}& 0.1 &  \multirow{3}{*}{-77.54 (6.71)}& -47.81 (14.97) & -43.77 (11.78)\\
&& 0.2 &  & -39.30 (10.36) & -36.84 (19.81) \\
&& 0.3 &  & -46.07 (8.76)& -29.68 (1.31) \\
\midrule
\multirow{9}{*}{3-shot}&\multirow{3}{*}{0.1}& 0.1 &  \multirow{3}{*}{-50.65 (14.91)}& -29.65 (1.44) & -32.21 (3.35)\\ 
&& 0.2 &  & -28.60 (1.77) & -29.16 (17.55) \\
&& 0.3 &  & -31.23 (7.92) & -21.18 (4.93)\\
\cmidrule{2-6}
&\multirow{3}{*}{0.2}& 0.1 &  \multirow{3}{*}{-68.85 (26.13)}& -38.09 (2.91) & -44.65 (5.83) \\
&& 0.2 &  & -36.37 (1.00) &  -37.91 (24.76)\\
&& 0.3 &  & -40.38 (10.23) & -27.37 (5.46)\\
\cmidrule{2-6}
&\multirow{3}{*}{0.3}& 0.1 &  \multirow{3}{*}{-76.14 (31.39)}& -42.30 (4.95) & -50.76 (8.92)\\
&& 0.2 &  & -39.24 (0.48) &  -41.68 (28.03)\\
&& 0.3 &  & -43.67 (12.04) &  -29.31 (5.96)\\
\midrule
\multirow{9}{*}{5-shot}&\multirow{3}{*}{0.1}& 0.1 &  \multirow{3}{*}{-39.61 (3.98)}& -30.90 (2.30) & -24.47 (4.00)\\ 
&& 0.2 &  & -28.06 (4.60) & -23.87 (7.43) \\
&& 0.3 &  & -30.77 (8.03) & -21.44 (5.19)\\
\cmidrule{2-6}
&\multirow{3}{*}{0.2}& 0.1 &  \multirow{3}{*}{-55.74 (5.77)}& -39.68 (3.69) & -31.74 (7.96)\\
&& 0.2 &  & -37.25 (6.83) & -29.69 (7.82) \\
&& 0.3 &  & -40.39 (9.17) & -29.50 (5.74)\\
\cmidrule{2-6}
&\multirow{3}{*}{0.3}& 0.1 &  \multirow{3}{*}{-61.55 (6.15)}& -43.68 (3.34) & -35.31 (10.24)\\
&& 0.2 &  & -40.47 (7.44) &  -32.56 (8.37)\\
&& 0.3 &  & -43.56 (8.56) &  -32.27 (5.43)\\
\bottomrule\bottomrule
\end{tabular}
\end{table}

\paragraph{Results on AntVel.} For experiments deployed in ant environment with target velocity (AntVel), the rewards of the experiments are defined as the difference between the current velocity of the agent and a velocity goal in the range of 0.0 to 3.0. To maximize the diversity of the initialized velocity goal, we sample 20 different velocity values at equal interval in this range for the 20 generated tasks.
It can be observed from the experimental results shown in Table~\ref{tab:0_shot_antvel} that our proposed *GiRL and GiRL both show significant improvement over MAML. 
In detail, the analysis of the results can be summarized as follows: 
1) The return values of both MAML and our *GiRL decrease as $test\_max$ increases from 0.1 to 0.3. However, *GiRL decreases more gently, indicating that our methods have more balanced performance in dealing with multiple tasks. 2) For fixed $test\_max$ and increased $train\_max$, GiRL's performance improves, which shows that policies trained using GiRL can learn more information when increasing distribution freedom. The standard deviation of GiRL decreases along with the rise of $train\_max$, which justifies that the extra information learned also promotes model robustness. However, *GiRL is more prominent when $train\_max$ equals to 0.2, comparing to 0.1 and 0.3, which shows that the initialization of each iteration makes the policy lack accumulation of the previous learning process, leading to insufficient learning of some tasks. 3) The return of MAML increases more obviously than our *GiRL when adding fine-tuning process from 0-shot to 5-shot before evaluation. This shows that the policies obtained by our algorithm achieve a closer performance to single-task learning on individual tasks comparing to MAML.
 
\begin{table}[ht]
\centering
\caption{Experiments on AntPos with 20 diverse tasks.}
\label{tab:0_shot_antpos}
\begin{tabular}{c|c|c||c|cc}
\toprule\toprule
Shots&Test max     & Train max & MAML & *GiRL (Ours)& GiRL (Ours) \\
\midrule
\multirow{9}{*}{0-shot}&\multirow{3}{*}{0.1}& 0.1 &  \multirow{3}{*}{185.03 (40.72)}& 209.28 (1.34) & 176.25 (29.80)\\ 
&& 0.2 &  & 211.27 (14.92) & 186.62 (48.99) \\
&& 0.3 &  & 209.86 (4.46) & 161.70 (37.96) \\
\cmidrule{2-6}
&\multirow{3}{*}{0.2}& 0.1 &  \multirow{3}{*}{160.55 (46.81)}&  192.17 (9.49) & 152.40 (23.81)\\
&& 0.2 &  & 195.34 (16.65) & 178.33 (50.34)\\
&& 0.3 &  & 198.59 (10.53) & 137.18 (32.42)\\
\cmidrule{2-6}
&\multirow{3}{*}{0.3}& 0.1 &  \multirow{3}{*}{149.32 (48.14)}& 186.01 (14.38) & 138.44 (18.34)\\
&& 0.2 &  & 189.87 (18.83) &  173.35 (52.05)\\
&& 0.3 &  & 193.52 (13.83) &  127.11 (27.01)\\
\midrule

\multirow{9}{*}{3-shot}&\multirow{3}{*}{0.1}& 0.1 &  \multirow{3}{*}{192.47 (21.02)}& 207.96 (10.39) & 163.13 (57.05)\\
&& 0.2 &  & 222.71 (2.85) & 184.87 (43.60) \\
&& 0.3 &  & 214.51 (6.29) & 178.26 (26.80)\\
\cmidrule{2-6}
&\multirow{3}{*}{0.2}& 0.1 &  \multirow{3}{*}{167.59 (24.58)}& 185.51 (8.06) &  129.35 (62.79) \\
&& 0.2 &  & 210.28 (3.55) &  166.69 (48.15)\\
&& 0.3 &  & 201.60 (13.04) & 144.82 (24.80)\\
\cmidrule{2-6}
&\multirow{3}{*}{0.3}& 0.1 &  \multirow{3}{*}{158.69 (26.75)}& 180.51 (5.60) & 114.94 (63.60)\\
&& 0.2 &  & 206.41 (4.04) &  159.56 (49.93)\\
&& 0.3 &  & 198.33 (13.76)& 128.43 (24.71) \\
\midrule
\multirow{9}{*}{5-shot}&\multirow{3}{*}{0.1}& 0.1 &  \multirow{3}{*}{180.18 (40.47)}& 221.82 (13.51) & 178.54 (64.28)\\ 
&& 0.2 &  & 222.36 (12.12) &  184.19 (52.49)\\
&& 0.3 &  & 218.45 (13.09) & 174.03 (42.38)\\
\cmidrule{2-6}
&\multirow{3}{*}{0.2}& 0.1 &  \multirow{3}{*}{153.70 (41.22)}& 205.62 (13.93) & 153.51 (68.96)\\
&& 0.2 &  & 206.87 (10.22) & 167.78 (49.47) \\
&& 0.3 &  & 204.15 (10.36) & 145.41 (30.90)\\
\cmidrule{2-6}
&\multirow{3}{*}{0.3}& 0.1 &  \multirow{3}{*}{143.51 (43.27)}& 199.94 (14.72) & 139.61 (70.94)\\
&& 0.2 &  & 199.11 (11.79) &  161.92 (49.39)\\
&& 0.3 &  & 197.32 (10.01) &  132.75 (25.48)\\
\bottomrule\bottomrule
\end{tabular}
\end{table}
\paragraph{Results on AntPos.} For experiments deployed in ant environment with target position (AntPos), the reward of the experiments is defined as the L1 distance between the position of the agent and the target position within the range of $[-3.0, 3.0]^2$. We sample 20 bisection points on a circle with center $[0, 0]$ and radius 2.0 as the positions for the 20 constructed tasks. Therefore, the initialization is largely diverse and can effectively eliminate random deviation. It can be observed from the experimental results shown in Table~\ref{tab:0_shot_antpos} that our *GiRL outperforms MAML under each training and testing setting.
Our GiRL is also competitive with MAML. Different from AntVel, the deep learning model trained on AntPos is sensitive to the initialization, so that initializing the policy in each iteration has great influence on the final results of our method. The accumulation learning in GiRL causes overtraining and subsequent overfitting of the policy network. We conjecture the cause of this overtraining is that when initializing with the trained policy in the previous loop, the policy cannot efficiently find a good policy against the updated distribution over task. We also observe the *GiRL is robust to the uncertainties of the $test\_max$, i.e., our methods consistently perform better when the $test\_max$ differs from $train\_max$.

To summarize, the experiments on both AntVel and AntPos consistently show that: i) our PSRO-GiRL methods consistently outperform MAML against the adversary who can manipulate the task distribution, even with few-shot learning during evaluation, and ii) our methods are more robust in terms of generalization, thus provide better framework for real life deployment of RL.

\section{Conclusions}
In this work, motivated by the intrinsic similarities of various learning schemes that investigate the generalization in RL, we propose a game-theoretic framework GiRL, where an adversary is introduced to manipulate the distribution over tasks, for this issue. PSRO is adapted for GiRL with MAML as the best-response oracle and R-PRD for meta-solver. Extensive experiments on MuJoCo environments demonstrate that our proposed methods achieve better performance over existing baselines.
\paragraph{Future Works.} GiRL provides a general framework and there are many future directions worth exploring: i) generalizing to infinite number of tasks, where this is the exact formulation in robust and adversarial reinforcement learning as each adversary's policy will determine a task, ii) generalizing to unseen tasks, where the test tasks may never seen by the agent during training, which corresponds to the transfer learning. To handle the first direction, we need to learn a compact representation of the infinite number of tasks, i.e., mapping the task to a latent space. While for the second direction, we may need to learn a model to generate the possible unseen tasks from the seen tasks during training. We will leverage the model-based reinforcement learning methods~\cite{schrittwieser2020mastering} to improve the generalization of reinforcement learning in the GiRL framework~\cite{anand2021procedural}. Other directions such as the perturbation noise during training, the tasks across domains can also be explored. 

\bibliography{generalization}

\newcommand{\etalchar}[1]{$^{#1}$}
\begin{thebibliography}{GDW{\etalchar{+}}19}

\bibitem[AWL{\etalchar{+}}21]{anand2021procedural}
Ankesh Anand, Jacob Walker, Yazhe Li, Eszter V{\'e}rtes, Julian Schrittwieser,
  Sherjil Ozair, Th{\'e}ophane Weber, and Jessica~B Hamrick.
\newblock Procedural generalization by planning with self-supervised world
  models.
\newblock {\em arXiv preprint arXiv:2111.01587}, 2021.

\bibitem[AWY{\etalchar{+}}21]{aung2021gan}
Aye Phyu~Phyu Aung, Xinrun Wang, Runsheng Yu, Bo~An, Senthilnath Jayavelu, and
  Xiaoli Li.
\newblock {DO-GAN}: A double oracle framework for generative adversarial
  networks.
\newblock {\em arXiv preprint arXiv:2102.08577}, 2021.

\bibitem[BBQ{\etalchar{+}}18]{barreto2018transfer}
Andre Barreto, Diana Borsa, John Quan, Tom Schaul, David Silver, Matteo Hessel,
  Daniel Mankowitz, Augustin Zidek, and Remi Munos.
\newblock Transfer in deep reinforcement learning using successor features and
  generalised policy improvement.
\newblock In {\em ICML}, pages 501--510, 2018.

\bibitem[BKLP12]{bosansky2012iterative}
B~Bosansky, Christopher Kiekintveld, Viliam Lisy, and Michal Pechoucek.
\newblock Iterative algorithm for solving two-player zero-sum extensive-form
  games with imperfect information.
\newblock In {\em ECAI}, pages 193--198, 2012.

\bibitem[BS18]{brown2018superhuman}
Noam Brown and Tuomas Sandholm.
\newblock Superhuman {AI} for heads-up no-limit poker: Libratus beats top
  professionals.
\newblock {\em Science}, 359(6374):418--424, 2018.

\bibitem[BTHK15]{bloembergen2015evolutionary}
Daan Bloembergen, Karl Tuyls, Daniel Hennes, and Michael Kaisers.
\newblock Evolutionary dynamics of multi-agent learning: A survey.
\newblock {\em Journal of Artificial Intelligence Research}, 53:659--697, 2015.

\bibitem[DGD{\etalchar{+}}17]{devin2017learning}
Coline Devin, Abhishek Gupta, Trevor Darrell, Pieter Abbeel, and Sergey Levine.
\newblock Learning modular neural network policies for multi-task and
  multi-robot transfer.
\newblock In {\em ICRA}, pages 2169--2176, 2017.

\bibitem[DGP09]{daskalakis2009complexity}
Constantinos Daskalakis, Paul~W Goldberg, and Christos~H Papadimitriou.
\newblock The complexity of computing a {Nash} equilibrium.
\newblock {\em SIAM Journal on Computing}, 39(1):195--259, 2009.

\bibitem[FAL17]{finn2017model}
Chelsea Finn, Pieter Abbeel, and Sergey Levine.
\newblock Model-agnostic meta-learning for fast adaptation of deep networks.
\newblock In {\em ICML}, pages 1126--1135, 2017.

\bibitem[GDW{\etalchar{+}}19]{gleave2019adversarial}
Adam Gleave, Michael Dennis, Cody Wild, Neel Kant, Sergey Levine, and Stuart
  Russell.
\newblock Adversarial policies: Attacking deep reinforcement learning.
\newblock In {\em ICLR}, 2019.

\bibitem[GRK{\etalchar{+}}21]{ghosh2021generalization}
Dibya Ghosh, Jad Rahme, Aviral Kumar, Amy Zhang, Ryan~P Adams, and Sergey
  Levine.
\newblock Why generalization in {RL} is difficult: Epistemic {POMDPs} and
  implicit partial observability.
\newblock {\em NeurIPS}, 2021.

\bibitem[HVD{\etalchar{+}}15]{hinton2015distilling}
Geoffrey Hinton, Oriol Vinyals, Jeff Dean, et~al.
\newblock Distilling the knowledge in a neural network.
\newblock {\em arXiv preprint arXiv:1503.02531}, 2(7), 2015.

\bibitem[HZAL18]{haarnoja2018soft}
Tuomas Haarnoja, Aurick Zhou, Pieter Abbeel, and Sergey Levine.
\newblock Soft actor-critic: Off-policy maximum entropy deep reinforcement
  learning with a stochastic actor.
\newblock In {\em ICML}, pages 1861--1870, 2018.

\bibitem[JKV{\etalchar{+}}11]{jain2011double}
Manish Jain, Dmytro Korzhyk, Ond{\v{r}}ej Van{\v{e}}k, Vincent Conitzer, Michal
  P{\v{e}}chou{\v{c}}ek, and Milind Tambe.
\newblock A double oracle algorithm for zero-sum security games on graphs.
\newblock In {\em AAMAS}, pages 327--334, 2011.

\bibitem[KZGR21]{kirk2021survey}
Robert Kirk, Amy Zhang, Edward Grefenstette, and Tim Rockt{\"a}schel.
\newblock A survey of generalisation in deep reinforcement learning.
\newblock {\em arXiv preprint arXiv:2111.09794}, 2021.

\bibitem[LZG{\etalchar{+}}17]{lanctot2017unified}
Marc Lanctot, Vinicius Zambaldi, Audr{\=u}nas Gruslys, Angeliki Lazaridou, Karl
  Tuyls, Julien P{\'e}rolat, David Silver, and Thore Graepel.
\newblock A unified game-theoretic approach to multiagent reinforcement
  learning.
\newblock In {\em NeurIPS}, pages 4193--4206, 2017.

\bibitem[MBM{\etalchar{+}}16]{mnih2016asynchronous}
Volodymyr Mnih, Adria~Puigdomenech Badia, Mehdi Mirza, Alex Graves, Timothy
  Lillicrap, Tim Harley, David Silver, and Koray Kavukcuoglu.
\newblock Asynchronous methods for deep reinforcement learning.
\newblock In {\em ICML}, pages 1928--1937, 2016.

\bibitem[MGB03]{mcmahan2003planning}
H~Brendan McMahan, Geoffrey~J Gordon, and Avrim Blum.
\newblock Planning in the presence of cost functions controlled by an
  adversary.
\newblock In {\em ICML}, pages 536--543, 2003.

\bibitem[MKS{\etalchar{+}}15]{mnih2015human}
Volodymyr Mnih, Koray Kavukcuoglu, David Silver, Andrei~A Rusu, Joel Veness,
  Marc~G Bellemare, Alex Graves, Martin Riedmiller, Andreas~K Fidjeland, Georg
  Ostrovski, et~al.
\newblock Human-level control through deep reinforcement learning.
\newblock {\em Nature}, 518(7540):529--533, 2015.

\bibitem[MOR{\etalchar{+}}20]{muller2020generalized}
P~Muller, S~Omidshafiei, M~Rowland, K~Tuyls, J~P{\'e}rolat, S~Liu, D~Hennes,
  L~Marris, M~Lanctot, E~Hughes, et~al.
\newblock A generalized training approach for multiagent learning.
\newblock In {\em ICLR}, 2020.

\bibitem[PBS16]{parisotto2016actor}
Emilio Parisotto, Lei~Jimmy Ba, and Ruslan Salakhutdinov.
\newblock Actor-mimic: Deep multitask and transfer reinforcement learning.
\newblock In {\em ICLR}, 2016.

\bibitem[PDSG17]{pinto2017robust}
Lerrel Pinto, James Davidson, Rahul Sukthankar, and Abhinav Gupta.
\newblock Robust adversarial reinforcement learning.
\newblock In {\em ICML}, pages 2817--2826, 2017.

\bibitem[PGM{\etalchar{+}}19]{paszke2019pytorch}
Adam Paszke, Sam Gross, Francisco Massa, Adam Lerer, James Bradbury, Gregory
  Chanan, Trevor Killeen, Zeming Lin, Natalia Gimelshein, Luca Antiga, et~al.
\newblock {PyTorch}: An imperative style, high-performance deep learning
  library.
\newblock In {\em NeurIPS}, pages 8026--8037, 2019.

\bibitem[PY09]{pan2009survey}
Sinno~Jialin Pan and Qiang Yang.
\newblock A survey on transfer learning.
\newblock {\em IEEE Transactions on Knowledge and Data Engineering},
  22(10):1345--1359, 2009.

\bibitem[RCG{\etalchar{+}}16]{rusu2016policy}
Andrei~A Rusu, Sergio~Gomez Colmenarejo, {\c{C}}aglar G{\"u}l{\c{c}}ehre,
  Guillaume Desjardins, James Kirkpatrick, Razvan Pascanu, Volodymyr Mnih,
  Koray Kavukcuoglu, and Raia Hadsell.
\newblock Policy distillation.
\newblock In {\em ICLR}, 2016.

\bibitem[RRBV19]{raghu2019rapid}
Aniruddh Raghu, Maithra Raghu, Samy Bengio, and Oriol Vinyals.
\newblock Rapid learning or feature reuse? towards understanding the
  effectiveness of {MAML}.
\newblock In {\em ICML}, 2019.

\bibitem[Rud17]{ruder2017overview}
Sebastian Ruder.
\newblock An overview of multi-task learning in deep neural networks.
\newblock {\em arXiv preprint arXiv:1706.05098}, 2017.

\bibitem[SAH{\etalchar{+}}20]{schrittwieser2020mastering}
Julian Schrittwieser, Ioannis Antonoglou, Thomas Hubert, Karen Simonyan,
  Laurent Sifre, Simon Schmitt, Arthur Guez, Edward Lockhart, Demis Hassabis,
  Thore Graepel, et~al.
\newblock Mastering {Atari}, {Go}, chess and shogi by planning with a learned
  model.
\newblock {\em Nature}, 588(7839):604--609, 2020.

\bibitem[SHM{\etalchar{+}}16]{silver2016mastering}
David Silver, Aja Huang, Chris~J Maddison, Arthur Guez, Laurent Sifre, George
  Van Den~Driessche, Julian Schrittwieser, Ioannis Antonoglou, Veda
  Panneershelvam, Marc Lanctot, et~al.
\newblock Mastering the game of {Go} with deep neural networks and tree search.
\newblock {\em Nature}, 529(7587):484--489, 2016.

\bibitem[SLA{\etalchar{+}}15]{schulman2015trust}
John Schulman, Sergey Levine, Pieter Abbeel, Michael Jordan, and Philipp
  Moritz.
\newblock Trust region policy optimization.
\newblock In {\em ICML}, pages 1889--1897, 2015.

\bibitem[SWD{\etalchar{+}}17]{schulman2017proximal}
John Schulman, Filip Wolski, Prafulla Dhariwal, Alec Radford, and Oleg Klimov.
\newblock Proximal policy optimization algorithms.
\newblock {\em arXiv preprint arXiv:1707.06347}, 2017.

\bibitem[TNT12]{tsai2012security}
Jason Tsai, Thanh~H Nguyen, and Milind Tambe.
\newblock Security games for controlling contagion.
\newblock In {\em AAAI}, pages 1464--1470, 2012.

\bibitem[VBC{\etalchar{+}}19]{vinyals2019grandmaster}
Oriol Vinyals, Igor Babuschkin, Wojciech~M Czarnecki, Micha{\"e}l Mathieu,
  Andrew Dudzik, Junyoung Chung, David~H Choi, Richard Powell, Timo Ewalds,
  Petko Georgiev, et~al.
\newblock Grandmaster level in {StarCraft II} using multi-agent reinforcement
  learning.
\newblock {\em Nature}, 575(7782):350--354, 2019.

\bibitem[VDP{\etalchar{+}}20]{vinitsky2020robust}
Eugene Vinitsky, Yuqing Du, Kanaad Parvate, Kathy Jang, Pieter Abbeel, and
  Alexandre Bayen.
\newblock Robust reinforcement learning using adversarial populations.
\newblock {\em arXiv preprint arXiv:2008.01825}, 2020.

\bibitem[WBB09]{waugh2009strategy}
Kevin Waugh, Nolan Bard, and Michael Bowling.
\newblock Strategy grafting in extensive games.
\newblock In {\em NeurIPS}, pages 2026--2034, 2009.

\bibitem[ZCBH20]{zhang2020robust}
Huan Zhang, Hongge Chen, Duane~S Boning, and Cho-Jui Hsieh.
\newblock Robust reinforcement learning on state observations with learned
  optimal adversary.
\newblock In {\em ICLR}, 2020.

\bibitem[ZHB20]{zhang2020stability}
Kaiqing Zhang, Bin Hu, and Tamer Basar.
\newblock On the stability and convergence of robust adversarial reinforcement
  learning: A case study on linear quadratic systems.
\newblock {\em NeurIPS}, 33:22056--22068, 2020.

\bibitem[ZLZ20]{zhu2020transfer}
Zhuangdi Zhu, Kaixiang Lin, and Jiayu Zhou.
\newblock Transfer learning in deep reinforcement learning: A survey.
\newblock {\em arXiv preprint arXiv:2009.07888}, 2020.

\end{thebibliography}
\bibliographystyle{alpha}

\end{document}